\documentclass[review]{elsarticle}

\usepackage{lineno,hyperref}
\modulolinenumbers[5]
\usepackage{graphicx}
\usepackage{amsmath}
\usepackage{amssymb}
\usepackage{multirow}
\usepackage{booktabs} 
\usepackage{algorithm}
\usepackage{algorithmicx}
\usepackage{algpseudocode}
\journal{Neurocomputing}

\begin{document}

\begin{frontmatter}

\title{ZstGAN: An Adversarial Approach for Unsupervised Zero-Shot Image-to-Image Translation}


\author[mymainaddress]{Jianxin Lin}
\ead{linjx@mail.ustc.edu.cn}

\author[mysecondaryaddress]{Yingce Xia}
\ead{Yingce.Xia@microsoft.com}

\author[mymainaddress]{Sen Liu}
\ead{elsen@iat.ustc.edu.cn}

\author[mymainaddress]{Shuxin Zhao}
\ead{sx2222@mail.ustc.edu.cn}

\author[mymainaddress]{Zhibo Chen\corref{mycorrespondingauthor}}
\cortext[mycorrespondingauthor]{Corresponding author}
\ead{chenzhibo@ustc.edu.cn}

\address[mymainaddress]{Department of Electronic Engineering and Information Science, University of Science and Technology of China, Hefei, Anhui, 230026, China}
\address[mysecondaryaddress]{Microsoft Research Asia, Beijing, 100080, China}

\begin{abstract}
Image-to-image translation models have shown remarkable ability on transferring images among different domains. Most of existing work follows the setting that the source domain and target domain keep the same at training and inference phases, which cannot be generalized to the scenarios for translating an image from an unseen domain to another unseen domain. In this work, we propose the Unsupervised Zero-Shot Image-to-image Translation (UZSIT) problem, which aims to learn a model that can translate samples from image domains that are not observed during training. Accordingly, we propose a framework called ZstGAN: By introducing an adversarial training scheme, ZstGAN learns to model each domain with  domain-specific feature distribution that is semantically consistent on vision and attribute modalities. Then the domain-invariant features are disentangled with an shared encoder for image generation. We carry out extensive experiments on CUB and FLO datasets, and the results demonstrate the effectiveness of proposed method on UZSIT task. Moreover, ZstGAN shows significant accuracy improvements over state-of-the-art zero-shot learning methods on CUB and FLO.
\end{abstract}


\end{frontmatter}

\begin{abstract}
	Image-to-image translation models have shown remarkable ability on transferring images among different domains. Most of existing work follows the setting that the source domain and target domain keep the same at training and inference phases, which cannot be generalized to the scenarios for translating an image from an unseen domain to another unseen domain. In this work, we propose the Unsupervised Zero-Shot Image-to-image Translation (UZSIT) problem, which aims to learn a model that can translate samples from image domains that are not observed during training. Accordingly, we propose a framework called ZstGAN: By introducing an adversarial training scheme, ZstGAN learns to model each domain with  domain-specific feature distribution that is semantically consistent on vision and attribute modalities. Then the domain-invariant features are disentangled with an shared encoder for image generation. We carry out extensive experiments on CUB and FLO datasets, and the results demonstrate the effectiveness of proposed method on UZSIT task. Moreover, ZstGAN shows significant accuracy improvements over state-of-the-art zero-shot learning methods on CUB and FLO.
\end{abstract}
\section{Introduction}
\begin{figure}[htb!]
	\centerline{\includegraphics[width=7.5cm]{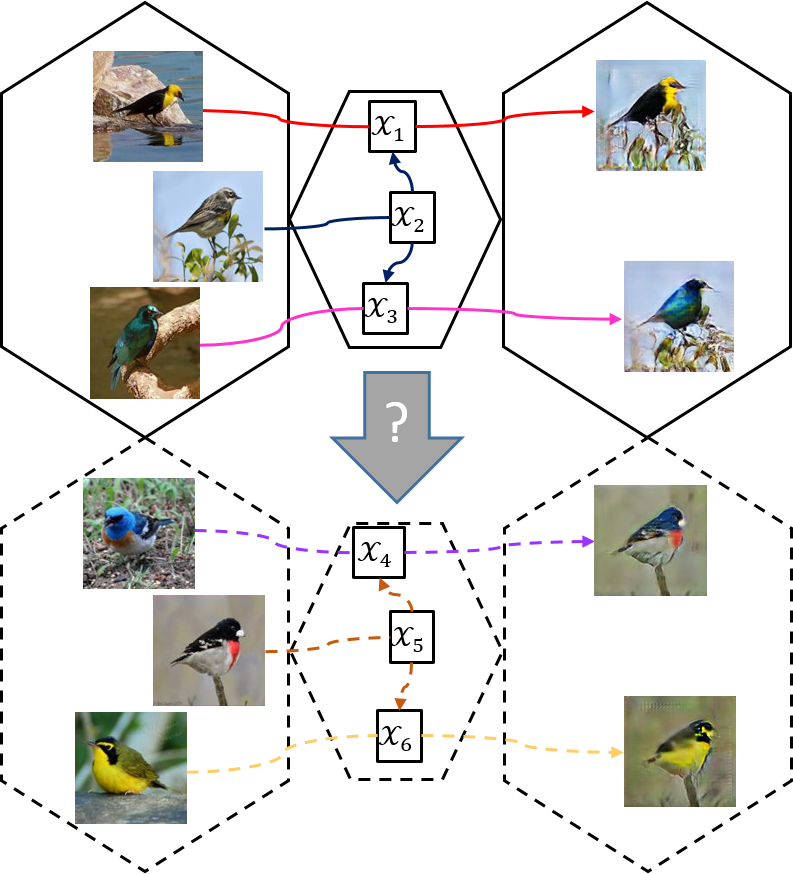}}
	\caption{Suppose that $\mathcal{X}_1$, $\mathcal{X}_2$ and $\mathcal{X}_3$ are seen domains for image translator training, and $\mathcal{X}_4$, $\mathcal{X}_5$ and $\mathcal{X}_6$ are unseen domains at inference phase. The aim of unsupervised zero-shot image-to-image translation is to translate images on unseen domains using translator trained on seen domains. }
	\centering
	\label{fig:zst_inllustration}
\end{figure}
Image-to-image translation tasks \cite{isola2016image,zhu2017unpaired}, which aim at learning mappings that can convert an image among different domains while preserving the main representations of the input images, have been widely investigated in recent years. Existing image-to-image translation usually works on the following setting: Given $M$ domains of interests, denoted as $\mathcal{X}_1,\mathcal{X}_2,\cdots\mathcal{X}_M$ where $M\ge2$, the objective is to learn mappings $f_{ij}:\mathcal{X}_i\mapsto\mathcal{X}_j$, where $i\ne j$. After obtaining these $f_{ij}$'s, we can achieve the translations among these $M$ domains. Specially, many models have been proposed for the setting $M=2$ like CycleGAN~\cite{zhu2017unpaired}, DiscoGAN~\cite{kim2017learning}, etc and for the setting $M>2$ like StarGAN~\cite{choi2017stargan}.

One limitation of existing models is, the $f_{ij}$'s can only achieve mappings among these given domains, without the generalization abilities to other unseen domains. That is, existing image-to-image translation models cannot translate an image from an unseen domain to another unseen domain. Take the bird translation shown in Figure \ref{fig:zst_inllustration} as an example. Assume a model $f$ is trained on domain $\mathcal{X}_1$, $\mathcal{X}_2$ and $\mathcal{X}_3$. Therefore, it is natural that $f$ can achieve translation among these three domains (see the upper half part of Figure~\ref{fig:zst_inllustration}), but $f$ cannot be applied to unseen domains $\mathcal{X}_4$, $\mathcal{X}_5$ and $\mathcal{X}_6$ (see the bottom half part of Figure~\ref{fig:zst_inllustration}). In practice, new image domains always come and it is impractical to train new translation models from scratch covering the new domains. Therefore we aim to generalize $f$ to unseen domains as shown in the bottom half part of Figure~\ref{fig:zst_inllustration}.

Zero-Shot Learning (ZSL) ~\cite{lampert2009learning,norouzi2013zero,long2018zero} aims to recognize objects whose instances might not have been seen during training. In order to generalize to unseen classes, a common assumption in zero-shot learning  is that some side-information about the classes is available, such as class attributes or textual descriptions, which provides semantic information about the classes.

As far as we can survey, there is no literature works on zero-shot learning for unsupervised image translation. To fulfill such a blank in image-to-image translation, we propose a new problem, \emph{unsupervised zero-shot image-to-image translation} (briefly, UZSIT), which aims to learn a model that can translate images from domains that are unseen in the training phase. In this problem, the term ``unsupervised'' means that we do not use any paired data from different domains for image translation process. In addition, the term ``zero-shot'' means that the category labels of image domains are disjoint in the training and testing phase. Compared to the standard ZSL, UZSIT is more challenging: (1) The target of image translation is more complex than classification, which not only requires us to generate representative features across seen and unseen domains but also generate reasonable translation images. (2) Unlike ZSL methods trained in a supervised way on seen domains, we do not have any paired data between any two domains. This requires us to learn the mappings in a fully unsupervised manner for both seen domains and unseen domains. Therefore, we devise a framework, called ZstGAN, for UZSIT problem. There are two key steps in ZstGAN.

\noindent(1) We model each seen/unseen domain using a domain-specific feature distribution constrained by semantic consistency. Specifically, a visual-to-semantic encoder, which transfers image input into hidden representations, and an attribute-to-semantic encoder, which transfers textual description input into hidden representations,  are introduced. They are jointly trained to extract domain-specific features from images and attributes respectively while preserving the same semantic information between these two modalities. The adversarial and classification losses are introduced to the two encoders to regularize training.

\noindent(2) We disentangle domain-invariant features from the domain-specific features and combine them to generate translation results, which is achieved by one adversarial learning loss and two reconstruction losses.

We work on two datasets commonly used in ZSL, Caltech-UCSD-Birds 200-2011 (CUB) \cite{WahCUB_200_2011} and Oxford Flowers (FLO) \cite{nilsback2008automated}, to verify the effectiveness of our method on UZSIT task. We also generalize our model to traditional ZSL tasks, and find that our model can achieve significant improvement over state-of-the-art ZSL methods on CUB and FLO datasets.

The remaining part is organized as follows: We present a brief over review of related works in Section \ref{sec:related work}. We detail the problem formulation of UZSIT and a description of our approach in Section \ref{sec:framework}.  The datasets and experimental results are reported in Section~\ref{sec:exps}. Finally, we summarize our work in the Section \ref{sec:conclusion}.

\section{Related Works}\label{sec:related work}
\textbf{Generative Adversarial Networks}
Image generation has been widely investigated in recent years. Most of works focus on modeling the natural image distribution. Generative Adversarial Network (GAN) \cite{goodfellow2014generative} was firstly proposed to generate images from random variables by a two-player minimax game: a generator G tries to create fake but plausible images, while a discriminator D is trained to distinguish difference between real and fake images. To address the stability issues in GAN, Wasserstein-GAN (WGAN) \cite{arjovsky2017wasserstein} was proposed to optimize an approximation of the Wasserstein distance. To further improve the vanishing and exploding gradient problems of WGAN, Gulrajani et al. \cite{gulrajani2017improved} proposed a WGAN-GP that uses gradient penalty instead of the weight clipping to enforce the Lipschitz constrain in WGAN. Mao et al. \cite{mao2017least} also proposed a LSGAN and found that optimizing the least square cost function is the same as optimizing a Pearson $\chi^2$ divergence. In this paper, we combine with WGAN-GP \cite{gulrajani2017improved} to generate domain-specific features and translation images.

\textbf{Image-to-Image Translation}
Recently, Isola et al. \cite{isola2016image} proposed a general conditional GAN (Pix2Pix) for a wide range of supervised image-to-image translation tasks, including label-to-street scene, aerial-to-map, day-to-night and so on. Discovering that image translation between two domains should obey the cycle consistent rule, DualGAN~\cite{Yi_2017_ICCV}, DiscoGAN~\cite{kim2017learning} and CycleGAN~\cite{zhu2017unpaired} were proposed to tackle the unpaired image translation problem by training two cross-domain translation models at the same time. However, CycleGANs lack the ability to control the translated results in the target domain and their results usually lack of diversity.  In order to control the translated results in the target domain and obtain more diverse outputs with a fixed input, works \cite{lin2018conditional,huang2018multimodal,lee2018diverse} divided the latent space whthin translation into domain-invariant and domain-specific portions. The different domains share the same domain-invariant latent space while each domain has different domain-specific latent spaces.   Choi et al. \cite{choi2017stargan} further proposed to perform image-to-image translations for multiple domains. Applying image-to-image translation to other computer vision tasks have also been explored in recent works \cite{ji2018saliency,zheng2019unpaired,guo2019gan}.  \cite{benaim2018one} and \cite{liu2019few} are most related to our work, as they propose a one-shot and few-shot image translation. In this work, we focus on a totally different setting as zero-shot image translation which learns to transfer images from unseen domains to other unseen domains.

\textbf{Zero-Shot Learning}
Zero-Shot Learning (ZSL) was first introduced by \cite{lampert2009learning}, where train and test classes are disjoint for object recognition. Traditional methods for ZSL are based on learning an embedding from the visual space to the semantic space. In the test period, the semantic vector of an unseen sample is extracted and the most likely class is predicted by nearest neighbor method \cite{wang2016relational,norouzi2013zero,socher2013zero}. Recent works on ZSL have widely explored the idea of generative models. Wang et al. \cite{wang2018zero} presented a deep generative model for ZSL based on VAE \cite{kingma2013auto}. Due to the rapidly developed GANs, other approaches used GANs to synthesize visual representations for the seen and unseen classes \cite{long2018zero,bucher2017generating}. However, the generate images usually lack sufficient quality to train a classifier for both the seen and unseen classes. Hence authors \cite{xian2018feature,felix2018multi} used GANs to synthesizes CNN features rather than image pixels conditioned on class-level semantic information. On the other hand, considering that ZSL is a domain shift problem, \cite{NIPS2013_5027,chao2016empirical} presented the Generalized ZSL (GZSL) that leverages both seen and unseen classes at test time.
\section{Framework}\label{sec:framework}
\begin{figure*}[htb!]
	\centerline{\includegraphics[width=11.5cm]{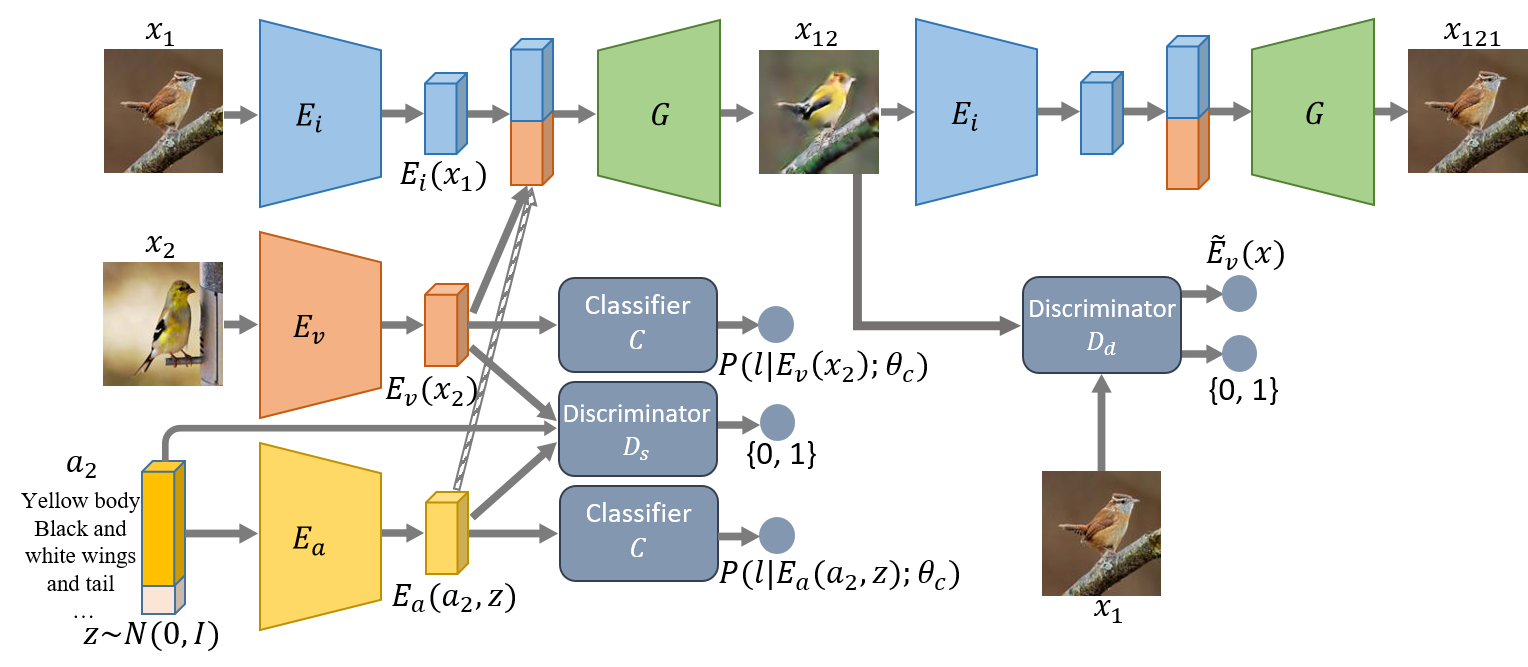}}
	\caption{The proposed ZstGAN framework. The dash line of $E_a(a_2,z)$ represents that $E_a(a_2,z)$ can be combined with $E_i(x_1)$ to generate translation result at inference phase. $\tilde{E}_v(x)$ represents domain-specific feature prediction of input to $D_d$. }
	\centering
	\label{fig:framework}
\end{figure*}

\subsection{Problem Formulation}\label{sec:problem}
We provide a mathematical formulation of UZSIT in this subsection.

Let $\mathcal{X}$ be the collection of images. Let $\mathcal{L}^s$ and $\mathcal{L}^u$ be two disjoint image categories, where $\mathcal{L}^s=\{l_1^s,l_2^s,\cdots,l^s_M\}$ and $\mathcal{L}^u=\{l_1^u,l_2^u,\cdots,l^u_N\}$, $M\ge2$, $N\ge1$ and $\mathcal{L}^s\cap\mathcal{L}^u=\emptyset$. For ease of reference, define $\mathcal{L}=\mathcal{L}^s\cup\mathcal{L}^u$. Let $\mathcal{A}$ denote the set of attributes or textual descriptions of images. Each sample can be represented by a $(x,l,a)\in\mathcal{X}\times\mathcal{L}\times\mathcal{A}$ where $x$ is a picture, $l$ is the corresponding label (e.g. a bird or a cat, etc) and $a$ is the attribute (e.g., the color, position, etc).  We have two different sets, a training set $\mathcal{S}=\{(x,l,a)|x\in\mathcal{X},l\in\mathcal{L}^s,a\in\mathcal{A}\}$ and a test set $\mathcal{U}=\{(x,l,a)|x\in\mathcal{X},l\in\mathcal{L}^u,a\in\mathcal{A}\}$.   

The objective of UZSIT is to train an image-to-image translation model $f$ on $\mathcal{S}$ without touching $\mathcal{U}$.  Then evaluating the obtained model $f$ on $\mathcal{U}$ without any further tuning. An assumption that $\mathcal{S}$ and $\mathcal{U}$ shares a common semantic space is required. Specifically,  while $\mathcal{S}$ and $\mathcal{U}$ have different category sets ($\mathcal{L}^s$ and $\mathcal{L}^u$), they are required to share the same image and attribute spaces ($\mathcal{X}$ and $\mathcal{A}$)  where semantic information is extracted from.

An implicit assumption in image-to-image translation is that an image contains two kinds of features~\cite{lin2018conditional,huang2018multimodal,lee2018diverse}: domain-invariant features $x^i\in\mathbb{R}^{d}$ and domains-specific features $x^s\in\mathbb{R}^{d}$ for any $x\in\mathcal{X}$, $d\in\mathbb{N}$. With an oracle image merge operator $\oplus$, $x=x^i\oplus x^s$.

In existing image-to-image translation models, the domains-specific features of different domains are usually extracted without depicting them in a common semantic space. So implicit relationship among different domains is omitted by this kind of features. In this paper, we argue that domains-specific features should be not only discriminative for different domains, but also representative to align different domains in a common semantic space. We will discuss how to learn the domains-specific features in the following subsection.

Depending on where the domains-specific features are extracted from, we devise two kinds of image translation problems at zero-shot testing phase.

\noindent(1) Vision-driven image translation:  $f_v:\mathcal{X}\oplus\mathcal{X}\mapsto\mathcal{X}$;

\noindent(2) Attribute-driven image translation: $f_a:\mathcal{X}\oplus\mathcal{A}\mapsto\mathcal{X}$.

In $f_v$,  the first image input $\mathcal{X}$ is used to provide domain-invariant features and the second image input $\mathcal{X}$ is used to specify domain-specific features. In $f_a$, the first image input $\mathcal{X}$ is used to provide domain-invariant features and the second attribute input $\mathcal{A}$ is used to specify domain-specific features.

\subsection{Architecture}

The architecture of our proposed ZstGAN is shown in Figure~\ref{fig:framework}. We use $\mathcal{N}(0,I)$ to denote a  Gaussian distribution. There are three encoders in our framework, domain-invariant encoder $E_i:\mathcal{X}\mapsto\mathbb{R}^d$, visual-to-semantic encoder $E_v:\mathcal{X}\mapsto\mathbb{R}^d$ and attribute-to-semantic encoder $E_a:\mathcal{A}\times\mathcal{N}(0,I)\to\mathbb{R}^d$, which work on extracting domain-invariant features, vision-based domain-specific features and attribute-based domain-specific features respectively. A decoder $G(\cdot,\cdot)$ is also needed to convert the hidden representations into natural images, where the first input is domain-invariant features and the second input is domain-specific features. That is, to generate an image, $f_v$ and $f_a$ works as follows:
\begin{equation}
	\begin{aligned}
		&f_v(x_1,x_2)=G(E_i(x_1),E_v(x_2))\\
		&f_a(x_1,a_2)=G(E_i(x_1),E_a(a_2,z)).
	\end{aligned}
	\label{eq:two_generation}
\end{equation}
where $x_1,x_2\in\mathcal{X}$, $a_2\in\mathcal{A}$, and $z$ is a noise vector sampled from $\mathcal{N}(0,I)$. We denote these two mappings with V-ZstGAN and A-ZstGAN respectively. In our configuration, we do not explicitly train $f_a$ in the training stage and it is naturally obtained by training $f_v$ with the following objective functions.

The objective functions are designed according to the following criteria:

\noindent(1) {\em Domain-specific features with semantic consistency}

Given a tuple $(x_2,l_2,a_2)\in\mathcal{S}$, the image $x_2$ and the attribute $a_2$ should share the same semantic representation. For such purpose, we utilize an adversarial training scheme which requires outputs of $E_{v}$ and $E_{a}$ to follow the same distribution conditioned on domain attributes. We need a domain-specific features discriminator $D_s$ which is used to distinguish outputs of $E_{v}$ and $E_{a}$. In detail, the adversarial training objective for $E_{v}$ and $E_{a}$ is:
\begin{equation}
	\begin{aligned}
		\ell_{\text{GAN,s}}  =& D_{s}(E_{v}(x_2),a_2)-D_{s}(E_{a}(a_2,z),a_2).
	\end{aligned}
	\label{eq:gan_loss_s}
\end{equation}

Using adversarial training can only ensure the distributions of the vision based domain-specific features and the attribute based domain-specific features to fit with each other. However, such features lack the ability to identify which domain the input images/attributes come from, causing meaninglessness of term ``domain-specific''. Thus, we require the $E_{v}(x_2)$ and $E_{a}(a_2,z)$ to be correctly classified by a classifier $C$. The classification loss is given as:
\begin{equation}
	\begin{aligned}
		& \ell_{\text{CLS,v}}  = -P(l=l_2|E_{v}(x_2);\theta_c),\; \\
		& \ell_{\text{CLS,a}}  = -P(l=l_2|E_{a}(a_2,z);\theta_c),
	\end{aligned}
	\label{eq:cls_losses_s}
\end{equation}
where $\theta_c$ is parameters of the classifier $C$.

\noindent(2) {\em Domain-invariant features disentanglement}

Given a domain-invariant encoder $E_{i}$ and a generator $G$ illustrated in Figure \ref{fig:framework} and another tuple $(x_1,l_1,a_1)\in\mathcal{S}$, we have domain-invariant features $E_{i}(x_1)$, domain-specific features $E_v(x_1)$. To translate image $x_1$ from domain $\mathcal{X}_1$ to domain $\mathcal{X}_2$, we can combine $E_{i}(x_1)$ and $E_v(x_2)$ to obtain $x_{12}=G(E_{i}(x_1),E_v(x_2))$. To ensure the translated result $x_{12}$ lie in the target domain $\mathcal{X}_2$ and in the real image domain, we introduce a domain discriminator $D_d$. $D_{d}^{\text{m}}$ of $D_d$ takes a real or fake image as input, and maximizes the mutual information between the target domain-specific features and the input as InfoGAN~\cite{chen2016infogan}.  Also, $D_{d}^{\text{g}}$ of $D_d$ outputs a probability of the input belonging to the real image domain. We illustrate the objective functions as below:
\begin{equation}
	\begin{aligned}
		\ell_{\text{MUT,r}}&=\Vert D_{d}^{\text{m}}(x_{1}) - E_{v}(x_1) \Vert_1,\\
		\ell_{\text{MUT,f}}& =\Vert D_{d}^{\text{m}}(x_{12}) - E_{v}(x_2) \Vert_1,
	\end{aligned}
	\label{eq:cls_loss_d}
\end{equation}
\begin{equation}
	\begin{aligned}
		\ell_{\text{GAN,d}}  = D_{d}^{\text{g}}(x_1)-D_{d}^{\text{g}}(x_{12}).
	\end{aligned}
	\label{eq:gan_loss_d}
\end{equation}

To ensure the disentanglement of domain-invariant features with domain-specific features, we introduce a self-reconstruction loss $\ell_{\text{REC,s}}$ and a cross-reconstruction loss $\ell_{\text{REC,c}}$. We can obtain the self-reconstructed image $x_{11}=G(E_{i}(x_1),E_v(x_1))$ and the cross-reconstructed image $x_{121}=G(E_i(x_{12}),E_v(x_1))$. The $\ell_{\text{REC,s}}$ is to minimize the L1 norm between $x_1$ and $x_{11}$:
\begin{equation}
	\begin{aligned}
		\ell_{\text{REC,s}}  = \Vert x_1 - x_{11} \Vert_1.
	\end{aligned}
	\label{eq:srec_loss}
\end{equation}

The $\ell_{\text{REC,c}}$ is to minimize the L1 norm between $x_1$ and $x_{121}$:
\begin{equation}
	\begin{aligned}
		\ell_{\text{REC,c}}  = \Vert x_1 - x_{121} \Vert_1.
	\end{aligned}
	\label{eq:rec_loss2}
\end{equation}

If $x_{11}$ optimally minimized $\ell_{\text{REC,s}}$ and $x_{12}$ optimally minimized $\ell_{\text{MUT,f}}$, we can find that the difference between $x_{11}$ and $x_{12}$, which are from two domains, only lies in the difference between  $E_{v}(x_1)$ and $E_{v}(x_2)$. Thus it implies that  $E_{v}(x_1)$ and $E_{v}(x_2)$ are domain-specific features that determine which domain image belongs to. On the other hand, if $x_{11}$ optimally minimized $\ell_{\text{REC,s}}$ and $x_{121}$ optimally minimized $\ell_{\text{REC,c}}$, we can find that the difference between $x_{11}$ and $x_{121}$, which are both the reconstruction images of the same $x_1$, only lies in the difference between  $E_i(x_{1})$ and $E_i(x_{12})$. Thus it further implies that $E_i(x_{1})$ and $E_i(x_{12})$ are domain-invariant features that maintain across different domains.

\noindent(3) {\em The overall training objective}

The overall objective for above mentioned encoders, discriminators, classifier and generator $E_v$, $E_a$, $C$, $D_s$, $E_i$, $G$ and $D_d$ is given by:

\begin{equation}
	\begin{aligned}
		& \ell^{\text{all}}_{E_v}= \ell_{\text{GAN,s}}+\lambda_c\ell_{\text{CLS,v}}, \ell^{\text{all}}_{E_a}= \ell_{\text{GAN,s}}+\lambda_c\ell_{\text{CLS,a}},\\
		& \ell^{\text{all}}_{C}= \ell_{\text{CLS,v}},  \ell^{\text{all}}_{D_s}= -\ell_{\text{GAN,s}},\\
		& \ell^{\text{all}}_{E_i}= \lambda_r\ell_{\text{REC,s}}+\lambda_r\ell_{\text{REC,c}} +\lambda_m\ell_{\text{MUT,f}} +\ell_{\text{GAN,d}},\\
		& \ell^{\text{all}}_{G}= \lambda_r\ell_{\text{REC,s}}+\lambda_r\ell_{\text{REC,c}} +\lambda_m\ell_{\text{MUT,f}} +\ell_{\text{GAN,d}},\\
		& \ell^{\text{all}}_{D_d}= \lambda_m\ell_{\text{MUT,r}} -\ell_{\text{GAN,d}},\\
		\label{eq:total_loss_g}
	\end{aligned}
\end{equation}
where $\lambda_{c}$, $\lambda_{r}$ and $\lambda_{m}$ are weights to achieve balance among different loss terms. Note that, unlike \cite{xian2018feature} that utilizes a pre-trained CNN feature extractor as a fixed visual-to-semantic encoder, our $E_v$ is updated with the adversarial and classification losses. The pre-trained CNN feature extractor restrict itself to adapt with specific domains and attributes, while our approach enables the $E_v$ to extract domain-specific features that are both discriminative for different domains and representative to align the visual images with attributes in a common semantic space. Experiment in the Section \ref{generalizing_zsl} also demonstrates that our approach significantly improves performance of \cite{xian2018feature}. The whole training procedure of ZstGAN is summarized in Algorithm \ref{alg:ZstGAN}.

\begin{algorithm}[!htpb]
	\caption{ZstGAN training process.}
	\label{alg:ZstGAN}
	\begin{algorithmic}[1]
		\State{\em Input:} $M$ seen domains $\mathcal{L}^s=\{l_1^s,l_2^s,\cdots,l^s_M\}$ and $N$ unseen domains $\mathcal{L}^u=\{l_1^u,l_2^u,\cdots,l^u_N\}$, $M\ge2$, $N\ge1$, a training set $\mathcal{S}=\{(x,l,a)|x\in\mathcal{X},l\in\mathcal{L}^s,a\in\mathcal{A}\}$ and a testing set $\mathcal{U}=\{(x,l,a)|x\in\mathcal{X},l\in\mathcal{L}^u,a\in\mathcal{A}\}$, learning rate $\eta=0.0001$.
		\State Randomly initialize the parameters $\Theta_{E_v}$ of $E_v$, $\Theta_{E_a}$ of $E_a$, $\Theta_C$ of $C$, $\Theta_{D_s}$ of $D_s$, $\Theta_{E_i}$ of $E_i$, $\Theta_{G}$ of $G$, and $\Theta_{D_d}$ of $D_d$.
		\State Randomly select a tuple $(x_2,l_2,a_2)\in\mathcal{S}$.
		\State Update the parameters $E_v$, $E_a$, $C$ and $D_s$ of as follows:
		\begin{equation}
		\begin{aligned}
		& \Theta_{E_v} \leftarrow \Theta_{E_v} - \eta\nabla_{\Theta_{E_v}}\ell^{\text{all}}_{E_v}, \\
		& \Theta_{E_a} \leftarrow \Theta_{E_a} - \eta\nabla_{\Theta_{E_a}}\ell^{\text{all}}_{E_a}, \\
		& \Theta_{C} \leftarrow \Theta_{C} - \eta\nabla_{\Theta_{C}}\ell^{\text{all}}_{C}, \\
		& \Theta_{D_s} \leftarrow \Theta_{D_s} - \eta\nabla_{\Theta_{D_s}}\ell^{\text{all}}_{D_s}, 
		\end{aligned}
		\end{equation}
		\State where $\ell^{\text{all}}_{E_v}$, $\ell^{\text{all}}_{E_a}$, $\ell^{\text{all}}_{C}$ and $\ell^{\text{all}}_{D_s}$ are defined in Eqn.~\eqref{eq:total_loss_g}.
		\State Repeat step 3 and step 5 until convergence.
		
		\State Randomly select a tuple $(x_2,l_2,a_2)\in\mathcal{S}$ and another tuple $(x_1,l_1,a_1)\in\mathcal{S}$. 
		\State Update the parameters of $E_i$, $G$ and $D_d$ as follows:
		\begin{equation}
		\begin{aligned}
		& \Theta_{E_i} \leftarrow \Theta_{E_i} - \eta\nabla_{\Theta_{E_i}}\ell^{\text{all}}_{E_i}, \\
		& \Theta_{G} \leftarrow \Theta_{G} - \eta\nabla_{\Theta_{G}}\ell^{\text{all}}_{G}, \\
		& \Theta_{D_d} \leftarrow \Theta_{D_d} - \eta\nabla_{\Theta_{D_d}}\ell^{\text{all}}_{D_d}, 
		\end{aligned}
		\end{equation}
		\State where $\ell^{\text{all}}_{E_i}$, $\ell^{\text{all}}_{G}$ and $\ell^{\text{all}}_{D_d}$ are defined in Eqn.~\eqref{eq:total_loss_g}.
		\State Repeat step 7 and step 9 until convergence.
		
	\end{algorithmic}
\end{algorithm}
\subsection{Implementation details}
For $E_v$, it consists of one $4\times 4$ stride $1$ convolution layer, six $4\times 4$ stride $2$ convolution layers, one $2 \times 2$ convolution layer for domain attribute-specific features output. One $1 \times 1$ convolution layer ($C$) is connected to $E_v$ for classification output.  For $E_a$, it shares a almost the same network architecture of $E_v$ except for the first convolution layer which takes attributes and noise as inputs. For discriminator $D_{s}$, it consists of two fully-connected layers as \cite{xian2018feature}. For domain discriminator $D_d$, we use PatchGANs \cite{isola2016image} that consists of six $4\times 4$ stride $2$ convolution layers, and two separated convolution layers for discrimination output and domain-specific feature prediction. For $E_i$, it has one $7\times 7$ stride $1$ convolution layer, two $4\times 4$ stride $2$ convolution layers and $3$ residual blocks \cite{he2016deep}. For generator $G$, it first processes the domain attribute-specific features through a fully-connected layer and adds it to domain-invariant features from encoder $E_i$. Then the combined feature is input to $3$ residual blocks, two $3\times 3$ stride $2$ deconvolution layers and one $7\times 7$ stride $1$ convolution layer. Specifically, we utilize the WGAN-GP \cite{gulrajani2017improved} to stablize our GAN training process. Our code is publicly available at \url{https://github.com/linjx-ustc1106/ZstGAN-PyTorch}.

For all experiments, we resize the images to $128\times 128$ resolution as inputs. The dimension of domain-specific features is set to $2048$. The dimension of $z$ is set to $312$. We set the weight parameters $\lambda_{c}=1$, $\lambda_{r}=1$ and $\lambda_{m}=10$.  We train our networks using Adam \cite{kingma2014adam} with learning rate of $0.0001$. For all experiments, we train models with a learning rate of $0.0001$ in the first $100000$ iterations and linearly decay the learning every $1000$ iteration.
\section{Experiments}\label{sec:exps}

\textbf{Datasets}
We conduct extensive quantitative and qualitative evaluations on Caltech-UCSD-Birds 200-2011 (CUB) \cite{WahCUB_200_2011} and Oxford Flowers (FLO) \cite{nilsback2008automated} which are commonly used in ZSL tasks. CUB contains $200$ bird species with $11,788$ images. We crop all images in CUB with bounding boxes given in  \cite{WahCUB_200_2011}. FLO contains $8,189$ images of flowers from $102$ different categories. For every image in CUB and FLO datasets, we extract 1024-dim character-based CNN-RNN \cite{reed2016learning} ($10$ captions are provided for each image) as the attribute set $\mathcal{A}$. We split each dataset into domain-disjoint train and test sets. CUB is split to $150$ train domains and $50$ unseen domains. Within $50$ unseen domains, $25\%$ data is used as test data; FLO is split to $82$ train domains and $20$ unseen domains. Within $20$ unseen domains, $25\%$ data is used as test data.

\begin{figure*}[htb!]
	\centerline{\includegraphics[width=12.0cm]{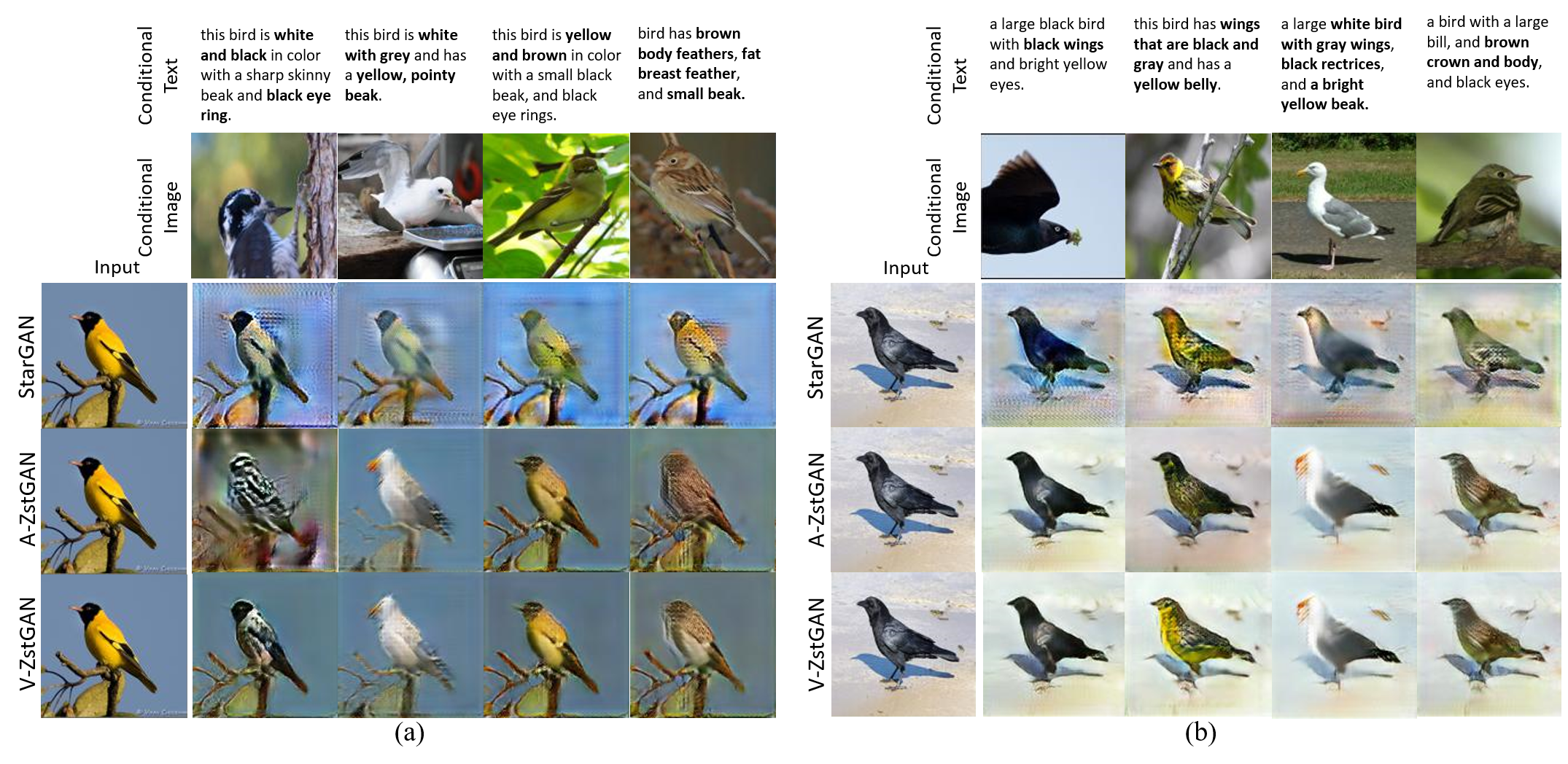}}
	\caption{Image translation results of StarGAN and our ZstGAN on unseen domains of CUB dataset. The first column is inputs $x_1$. The first row is $a_2$ for A-ZstGAN, and the second row is $x_2$ for V-ZstGAN and examples of target domain for StarGAN. Other images are the translation results. Note that StarGAN is trained on unseen domains of CUB. Key attributes contained in our translation results are in bold.}
	\centering
	\label{fig:comparison_bird}
\end{figure*}
\begin{figure*}[htb!]
	\centerline{\includegraphics[width=12.0cm]{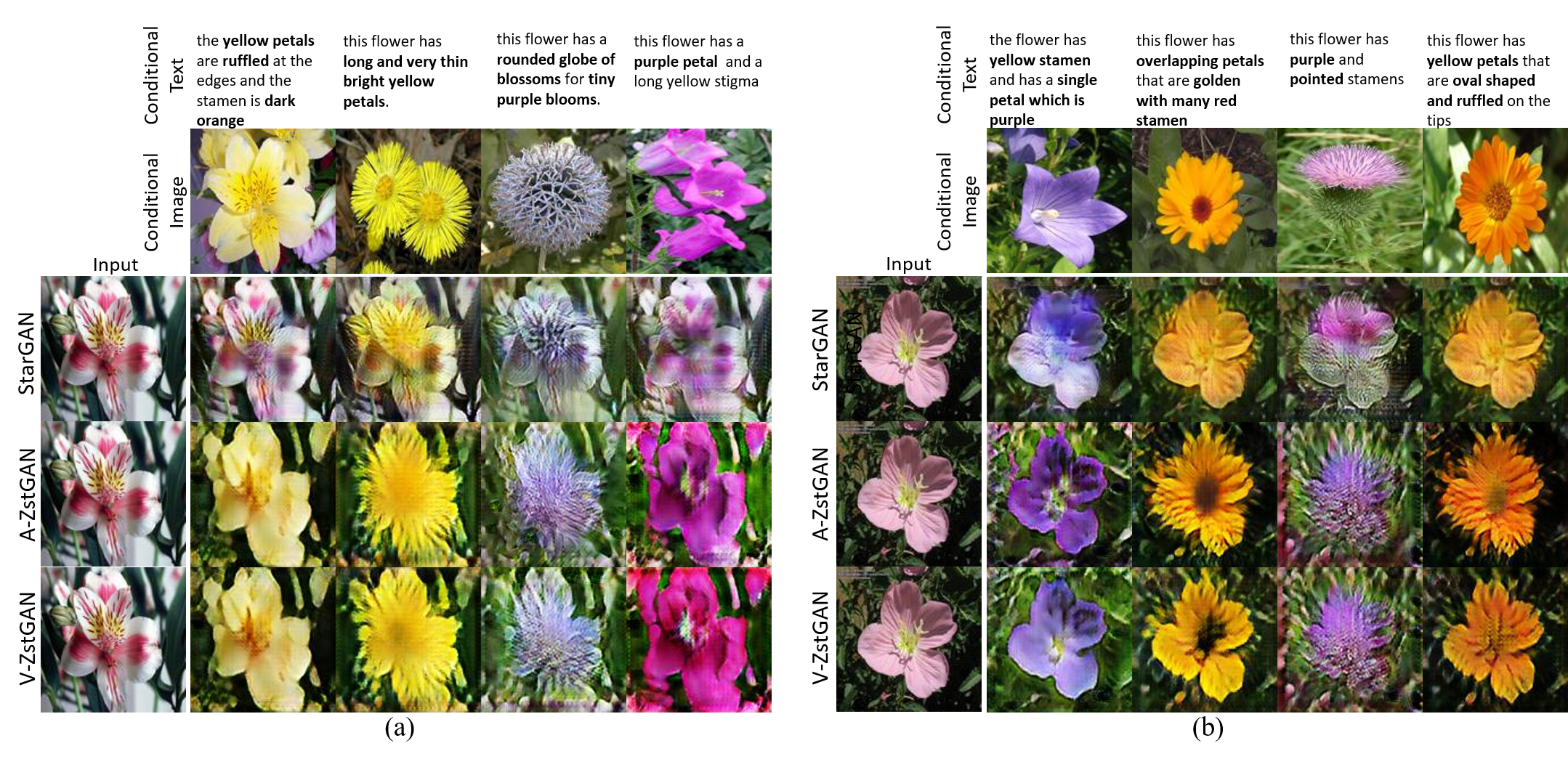}}
	\caption{Image translation results of StarGAN and our ZstGAN on unseen domains of FLO dataset. Note that StarGAN is trained on unseen domains of FLO. Key attributes contained in our translation results are in bold.}
	\centering
	\label{fig:comparison_flower}
\end{figure*}
\begin{table}[t]
	\caption{Top-1 and top-5 classification accuracy (\%) for translation results of StarGAN  and our ZstGAN on unseen domains of CUB and FLO datasets.}
	\label{table:comparison}
	\vskip 0.15in
	\begin{center}
		\begin{small}
			\begin{tabular}{lccccr}
				\toprule
				Dataset & StarGAN & A-ZstGAN & V-ZstGAN \\
				\midrule
				CUB(Top-1)    &15.24  &19.63 & \textbf{24.04} \\
				CUB(Top-5)    &43.32 &51.87 & \textbf{58.27} \\
				FLO(Top-1) &27.29 & 28.94 & \textbf{34.06}\\
				FLO(Top-5) &67.91 &  70.86& \textbf{75.12}\\
				\bottomrule
			\end{tabular}
		\end{small}
	\end{center}
	\vskip -0.1in
\end{table}
\begin{table}[t]
	\caption{FID scores for translation results of StarGAN  and our ZstGAN on unseen domains of CUB and FLO datasets.}
	\label{table:comparison_fid}
	\vskip 0.15in
	\begin{center}
		\begin{small}
			\begin{tabular}{lccccr}
				\toprule
				Dataset & StarGAN & A-ZstGAN & V-ZstGAN \\
				\midrule
				CUB    &99.95  &\textbf{81.98} & 82.35 \\
				FLO &110.2 & 96.46 & \textbf{94.87}\\
				\bottomrule
			\end{tabular}
		\end{small}
	\end{center}
	\vskip -0.1in
\end{table}
\subsection{Zero-Shot Image Translation Comparison}

Since there is no previous work on UZSIT problem, we compare with our model with StarGAN~\cite{choi2017stargan} that can be viewed as an unsupervised many-shot image-to-image translation model which is trained with data of unseen domains. We train StarGAN with data of total $200$ domains on CUB dataset, and with data of total $102$ domains on FLO dataset.

The translation results of StarGAN and our model on CUB and FLO are shown in Figure~\ref{fig:comparison_bird} and Figure~\ref{fig:comparison_flower}. We can find that although our ZstGAN is trained without data of unseen domains and StarGAN is trained with data of unseen domains, our ZstGAN shows even better translation quality with StarGAN in both CUB and FLO. For example, in the forth column of Figure~\ref{fig:comparison_bird}(b), our V-ZstGAN and A-ZstGAN accurately transfers the attributes of gray wings, black rectrices and bright yellow beak to the translation results, while StarGAN only shows little yellow and gray color without accurate position in the translation result. In the second column of Figure~\ref{fig:comparison_flower}(a), both A-ZstGAN and V-ZstGAN successfully transfer the ``long and very thin bright yellow petals'' description to the translation results, while StarGAN fails to change the shape of the original flower.  Such results are mainly due to the design of StarGAN that simply uses domain codes as domain-specific features, which make it difficult to align different domains with a common semantic space.  We can also see that translation results of A-ZstGAN highly correlate with V-ZstGAN's, which verifies the effectiveness of adversarial learning for vision based and attribute based domain-specific features alignment.

For quantitative evaluation, we translate source images from a random unseen domain to a random unseen domain in each test minibatch, and report the top-1 and top-5 classification accuracy of translated images of StarGAN and our model in Table \ref{table:comparison}, and Frechet Inception Distance (FID) \cite{NIPS2017_7240} scores in Table \ref{table:comparison_fid}. We can observe that the quantitative results are consistent with results in Figure~\ref{fig:comparison_bird} and Figure~\ref{fig:comparison_flower}, where our ZstGAN achieves better classification accuracy and FID scores than StarGAN.

\subsection{Generalizing to ZSL and GZSL}\label{generalizing_zsl}
The domain-specific features outputted from visual-to-semantic encoder $E_v$ and attribute-to-semantic encoder $E_a$ in our ZstGAN can also be used for ZSL and Generalized ZSL (GZSL) problems~\cite{NIPS2013_5027,chao2016empirical}, where in GZSL setting the seen domains can also be leveraged for testing. Specifically, we train two additional softmax classifiers that use generated domain-specific features from $E_a$ and corresponding labels as \cite{xian2018feature} for ZSL and GZSL testing respectively. In ZSL setting, only average per-class top-1 accuracy on unseen domains is computed. In GZSL setting, we compute average per-class top-1 accuracy on unseen domains (denoted as \textbf{U}), average per-class top-1 accuracy on seen domains (denoted as \textbf{S}) and their harmonic mean, i.e., $\textbf{H} =2 \times(\textbf{U}\times \textbf{S})/(\textbf{U}+ \textbf{S})$.

We compare our ZstGAN with three state-of-the-art ZSL and GZSL methods, e.g., SJE \cite{akata2015evaluation}, ESZSL \cite{romera2015embarrassingly} and f-CLSWGAN \cite{xian2018feature}. The ZSL results on CUB and FLO are shown in Table \ref{table:zsl_comparison}. The GZSL results on CUB and FLO are shown in Table \ref{table:gzsl_comparison}. The experiments clearly demonstrate the advantage of our ZstGAN for GZSL and ZSL since it achieves the best top-1 accuracy results in all the results, with improvements from $8\%$ to more than $21.1\%$. While our modification on f-CLSWGAN is not difficult to implement, our intuition is sound from the aspect of image-to-image translation and the improvement is rather significant. We also find that the classification accuracy of our model for ZSL is higher than the results for UZSIT in Table \ref{table:comparison}, this is because UZSIT is more challenging than ZSL since UZSIT needs to generate images that should properly fuse the domain-specific features with domain-invariant features to look like real target images.

\begin{figure*}[htb!]
	\centerline{\includegraphics[width=12cm]{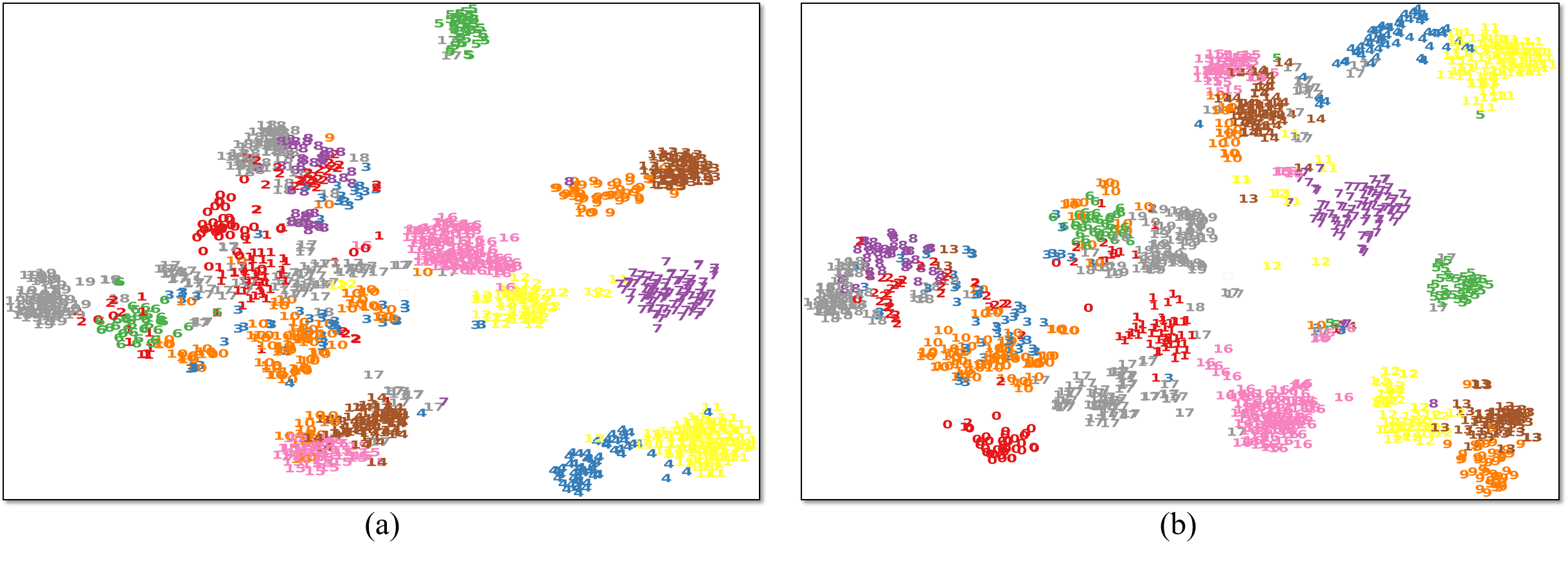}}
	\caption{t-SNE visualization of domain-specific features on FLO's unseen domains. (a) Domain-specific features extracted by $E_v$. (b) Domain-specific features extracted by $E_a$. The different colors and corresponding numbers indicate data of different unseen domains.}
	\centering
	\label{fig:visualization}
\end{figure*}

We also show the t-SNE \cite{maaten2008visualizing} visualization of domain-specific features extracted by $E_v$ and $E_a$ on unseen domains of FLO in Figure \ref{fig:visualization}. We can observe that: (1) Both Figure \ref{fig:visualization}(a) and Figure \ref{fig:visualization}(b) show clear clusters for different domains, which indicates that the $E_v$ and $E_a$ indeed learn to generalize to unseen domains; (2) Patterns of domain-specific features extracted by $E_v$ and $E_a$ are highly consistent to each other. For example, the samples of the 5th domain (green color) in Figure \ref{fig:visualization}(a) are mixed in some samples of the 17th domains, and the same phenomenon is observed in Figure \ref{fig:visualization}(b). Such result indicates that $E_v$ and $E_a$ indeed learn to mapping the visual images and attributes to the same semantic space.

\begin{table}[t]
	\caption{Top-1 accuracy results (\%) of ZSL on the unseen domains of CUB and FLO datasets.}
	\label{table:zsl_comparison}
	\vskip 0.15in
	\begin{center}
		\begin{small}
			\begin{tabular}{lccccr}
				\toprule
				Dataset & SJE & ESZSL & f-CLSWGAN & Ours \\
				\midrule
				CUB    & 53.9 & 53.9 & 57.3& \textbf{66.3}\\
				FLO & 53.4 & 51.0& 67.2& \textbf{70.7}\\
				\bottomrule
			\end{tabular}
		\end{small}
	\end{center}
	\vskip -0.1in
\end{table}

\begin{table}[t]
	\caption{Top-1 accuracy results (\%) of GZSL on the unseen domains (\textbf{U}), the seen domains (\textbf{S}) and the harmonic mean (\textbf{H}).}
	\label{table:gzsl_comparison}
	\vskip 0.15in
	\begin{center}
		\begin{small}
			\begin{tabular}{lccccr}
				\toprule
				Dataset &  & SJE & ESZSL & f-CLSWGAN & Ours \\
				\midrule
				\multirow{3}{*}{CUB}  &\textbf{U}   & 23.5 & 12.6 & 43.7& \textbf{61.5}\\
				&\textbf{S}   & 59.2 & 63.8 & 57.7& \textbf{83.5}\\
				&\textbf{H}   & 33.6 & 21.0 & 49.7& \textbf{70.8}\\
				\midrule
				\multirow{3}{*}{FLO} &\textbf{U}   & 13.9 & 11.4 & 59.0& \textbf{67.0}\\
				&\textbf{S}   & 47.6 & 56.8 & 73.8& \textbf{92.1}\\
				&\textbf{H}   & 21.5 & 19.0 & 65.6& \textbf{77.6}\\
				\bottomrule
			\end{tabular}
		\end{small}
	\end{center}
	\vskip -0.1in
\end{table}
\subsection{Analyzing Different Influence Factors of ZstGAN}
\subsubsection{Influence of losses}
We qualitatively compare V-ZstGAN with five variants of V-ZstGAN that ablate classification loss $\{\ell_{\text{CLS,v}}, \ell_{\text{CLS,a}}\}$, adversarial
loss $\ell_{\text{GAN,s}}$, reconstruction loss $\{\ell_{\text{REC,s}}, \ell_{\text{REC,c}}\}$, adversarial loss $\ell_{\text{GAN,d}}$, and information maximizing term $\{\ell_{\text{MUT,r}}, \ell_{\text{MUT,f}}\}$. As shown in Table \ref{table:ablation_study}, we can observe that ablating any specific loss will cause a big accuracy drop against original model, which verifies the necessity of the proposed training scheme.

\begin{table}[htb!]
	\caption{Ablation study of V-ZstGAN on on unseen domains of FLO dataset.}
	\label{table:ablation_study}
	\begin{center}
		\begin{tabular}{c c c c c}
			\toprule
			&Top-1 (\%)  & Top-5 (\%)     \\
			\midrule
			V-ZstGAN w/o $\{\ell_{\text{CLS,v}}, \ell_{\text{CLS,a}}\}$ &10.99 & 31.31      \\
			V-ZstGAN w/o $\ell_{\text{GAN,s}}$ &16.95 & 52.58\\
			V-ZstGAN w/o $\{\ell_{\text{REC,s}}, \ell_{\text{REC,c}}\}$ &18.86 & 57.45    \\
			V-ZstGAN w/o $\ell_{\text{GAN,d}}$ &12.32 & 36.59    \\
			V-ZstGAN w/o $\{\ell_{\text{MUT,r}}, \ell_{\text{MUT,f}}\}$ &8.24 & 25.63    \\
			V-ZstGAN &\textbf{34.06} & \textbf{75.12}\\
			\bottomrule
		\end{tabular}
	\end{center}
\end{table}

\subsubsection{Influence of $M$ Seen Domains}

To investigate how the number of seen domains influences the performance of zero-shot image translation on unseen domains. We train ZstGAN with different $M$ seen domains on FLO and show the classification accuracy results on unseen domains in Table \ref{table:M_on_FLO}. As we can see, with the decrease of $M$, the  classification accuracy of translation results also decreases. Such results are not surprising since the image translation on unseen domains is based on the semantic representation of seen domains. If semantic representation learned from seen domains is not adequate to represent semantic information of unseen domains, translation model may fail to translate image to the target domain.

\begin{table}[t]
	\centering
	\caption{Top-1 classification accuracy (\%) for translation results of our ZstGAN trained on different $M$ seen domain of FLO dataset.}
	\label{table:M_on_FLO}
	\begin{center}
		\begin{small}
			\begin{tabular}{lccccr}
				\toprule
				& $M$=20 & $M$=40 & $M$=60 & $M$=82\\
				\midrule
				A-ZstGAN &17.07 &17.13 &18.39 & \textbf{28.94}\\
				V-ZstGAN    &17.13 & 20.01& 23.31 &\textbf{34.06}\\
				\bottomrule
			\end{tabular}
		\end{small}
	\end{center}
\end{table}
\subsection{Interpolation}

\begin{figure}[t]
	\centerline{\includegraphics[width=8.5cm]{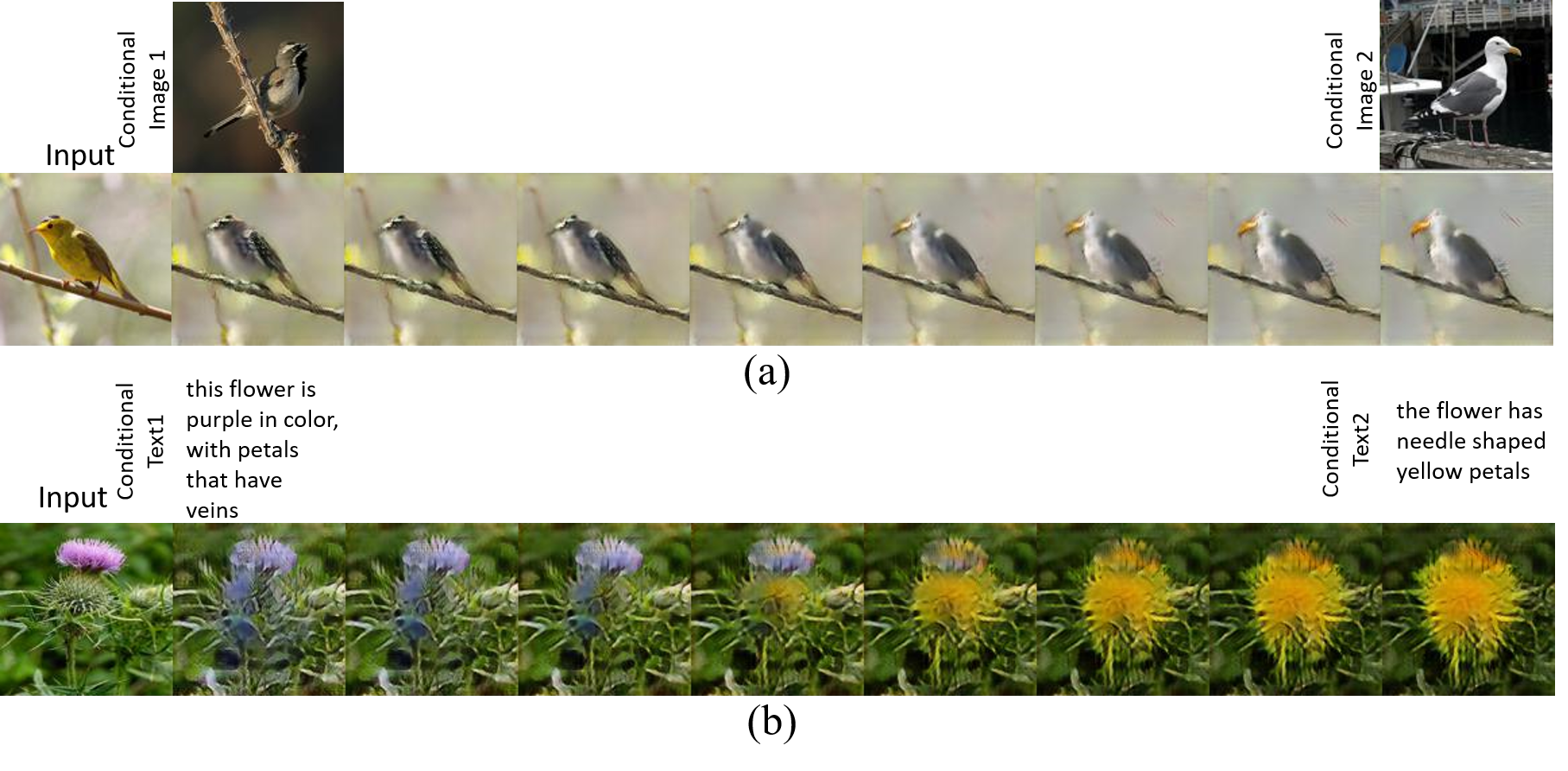}}
	\caption{Domain-specific features interpolations results on unseen domains. The translation results are generated by combining input's domain-invariant features with domain-specific features interpolated linearly from left conditional input's  to right conditional input's. (a) The interpolated domain-specific features are from conditional images. (b) The interpolated domain-specific features are from conditional texts. }
	\centering
	\label{fig:interpolation}
\end{figure}

To verify that the generality of our model is not only limited on the unseen domains given by specific datasets, we interpolate domain-specific features generated by images or texts from unseen domains for image translation. Specifically, given two conditional images, we linearly interpolate between their domain-specific features and combine the interpolated domain-specific features with domain-invariant features of input. Similar operation is used for conditional texts interpolation. The results of domain-specific features interpolations are shown in Figure~\ref{fig:interpolation}. We observe that our model can produce continuous translations through variation of  domain-specific features from both images and texts. This indicates that (1) our model indeed learns to generalize to unseen domains which are not only discrete ones given by specific datasets but also can be a continuous space covering the whole semantic representations; (2) our model learns to disentangle the domain-specific features and domain-invariant features since the domain-invariant features, such as leaves in the background, almost keep unchanged for different domain-specific features.

\section{Conclusions}\label{sec:conclusion}
In this paper, we propose an Unsupervised Zero-Shot Image-to-image Translation (UZSIT) problem, which aims to generalize image translation models from seen domains to unseen domains. Accordingly, we proposed a ZstGAN to this end. The ZstGAN models each seen/unseen domain using a domain-specific feature distribution conditioned on domain attributes, disentangles domain-invariant features from domain-specific features and combines them for image generation. Experiments show that our ZstGAN can effectively tackle UZSIT on CUB and FLO datasets. In addition, we show that ZstGAN can achieve much better performance than state-of-the-art ZSL and GZSL methods on CUB and FLO datasets.

For future work, there are many interesting directions.  First, it is interesting to design better models with better understanding of UZSIT. Second, achieving zero-shot translation without attributes is also valuable. Third, we may generalize the UZSIT to other relevant fields, such as domain adaptation and neural machine translation. 
\section*{Acknowledgments}
This work was supported in part by NSFC under Grant 61571413, 61632001.

\bibliography{Bibliography-File}

\begin{thebibliography}{10}
\expandafter\ifx\csname url\endcsname\relax
  \def\url#1{\texttt{#1}}\fi
\expandafter\ifx\csname urlprefix\endcsname\relax\def\urlprefix{URL }\fi
\expandafter\ifx\csname href\endcsname\relax
  \def\href#1#2{#2} \def\path#1{#1}\fi

\bibitem{isola2016image}
P.~{Isola}, J.~{Zhu}, T.~{Zhou}, A.~A. {Efros}, Image-to-image translation with
  conditional adversarial networks, in: 2017 IEEE Conference on Computer Vision
  and Pattern Recognition (CVPR), 2017, pp. 5967--5976.
\newblock \href {http://dx.doi.org/10.1109/CVPR.2017.632}
  {\path{doi:10.1109/CVPR.2017.632}}.

\bibitem{zhu2017unpaired}
J.-Y. Zhu, T.~Park, P.~Isola, A.~A. Efros, Unpaired image-to-image translation
  using cycle-consistent adversarial networks, in: The IEEE International
  Conference on Computer Vision (ICCV), 2017.

\bibitem{kim2017learning}
T.~Kim, M.~Cha, H.~Kim, J.~K. Lee, J.~Kim, Learning to discover cross-domain
  relations with generative adversarial networks, in: Proceedings of the 34th
  International Conference on Machine Learning, 2017, pp. 1857--1865.

\bibitem{choi2017stargan}
Y.~Choi, M.~Choi, M.~Kim, J.-W. Ha, S.~Kim, J.~Choo, Stargan: Unified
  generative adversarial networks for multi-domain image-to-image translation,
  in: The IEEE Conference on Computer Vision and Pattern Recognition (CVPR),
  2018.

\bibitem{lampert2009learning}
C.~H. Lampert, H.~Nickisch, S.~Harmeling, Learning to detect unseen object
  classes by between-class attribute transfer, in: Computer Vision and Pattern
  Recognition, 2009. CVPR 2009. IEEE Conference on, IEEE, 2009, pp. 951--958.

\bibitem{norouzi2013zero}
M.~Norouzi, T.~Mikolov, S.~Bengio, Y.~Singer, J.~Shlens, A.~Frome, G.~S.
  Corrado, J.~Dean, Zero-shot learning by convex combination of semantic
  embeddings, arXiv preprint arXiv:1312.5650.

\bibitem{long2018zero}
Y.~Long, L.~Liu, F.~Shen, L.~Shao, X.~Li, Zero-shot learning using synthesised
  unseen visual data with diffusion regularisation, IEEE transactions on
  pattern analysis and machine intelligence 40~(10) (2018) 2498--2512.

\bibitem{WahCUB_200_2011}
C.~Wah, S.~Branson, P.~Welinder, P.~Perona, S.~Belongie, {The Caltech-UCSD
  Birds-200-2011 Dataset}, Tech. Rep. CNS-TR-2011-001, California Institute of
  Technology (2011).

\bibitem{nilsback2008automated}
M.-E. Nilsback, A.~Zisserman, Automated flower classification over a large
  number of classes, in: Computer Vision, Graphics \& Image Processing, 2008.
  ICVGIP'08. Sixth Indian Conference on, IEEE, 2008, pp. 722--729.

\bibitem{goodfellow2014generative}
I.~Goodfellow, J.~Pouget-Abadie, M.~Mirza, B.~Xu, D.~Warde-Farley, S.~Ozair,
  A.~Courville, Y.~Bengio, Generative adversarial nets, in: Advances in neural
  information processing systems, 2014, pp. 2672--2680.

\bibitem{arjovsky2017wasserstein}
M.~Arjovsky, S.~Chintala, L.~Bottou, Wasserstein generative adversarial
  networks, in: International Conference on Machine Learning, 2017, pp.
  214--223.

\bibitem{gulrajani2017improved}
I.~Gulrajani, F.~Ahmed, M.~Arjovsky, V.~Dumoulin, A.~C. Courville, Improved
  training of wasserstein gans, in: Advances in Neural Information Processing
  Systems, 2017, pp. 5767--5777.

\bibitem{mao2017least}
X.~Mao, Q.~Li, H.~Xie, R.~Y. Lau, Z.~Wang, S.~Paul~Smolley, Least squares
  generative adversarial networks, in: Proceedings of the IEEE International
  Conference on Computer Vision, 2017, pp. 2794--2802.

\bibitem{Yi_2017_ICCV}
Z.~Yi, H.~Zhang, P.~Tan, M.~Gong, Dualgan: Unsupervised dual learning for
  image-to-image translation, in: The IEEE International Conference on Computer
  Vision (ICCV), 2017.

\bibitem{lin2018conditional}
J.~Lin, Y.~Xia, T.~Qin, Z.~Chen, T.-Y. Liu, Conditional image-to-image
  translation, in: The IEEE Conference on Computer Vision and Pattern
  Recognition (CVPR)(July 2018), 2018, pp. 5524--5532.

\bibitem{huang2018multimodal}
X.~Huang, M.-Y. Liu, S.~Belongie, J.~Kautz, Multimodal unsupervised
  image-to-image translation, in: ECCV, 2018.

\bibitem{lee2018diverse}
H.-Y. Lee, H.-Y. Tseng, J.-B. Huang, M.~K. Singh, M.-H. Yang, Diverse
  image-to-image translation via disentangled representations, in: European
  Conference on Computer Vision, 2018.

\bibitem{ji2018saliency}
Y.~Ji, H.~Zhang, Q.~J. Wu, Saliency detection via conditional adversarial
  image-to-image network, Neurocomputing 316 (2018) 357--368.

\bibitem{zheng2019unpaired}
Z.~Zheng, C.~Wang, Z.~Yu, N.~Wang, H.~Zheng, B.~Zheng, Unpaired
  photo-to-caricature translation on faces in the wild, Neurocomputing 355
  (2019) 71--81.

\bibitem{guo2019gan}
X.~Guo, Z.~Wang, Q.~Yang, W.~Lv, X.~Liu, Q.~Wu, J.~Huang, Gan-based
  virtual-to-real image translation for urban scene semantic segmentation,
  Neurocomputing.

\bibitem{benaim2018one}
S.~Benaim, L.~Wolf, One-shot unsupervised cross domain translation, arXiv
  preprint arXiv:1806.06029.

\bibitem{liu2019few}
M.-Y. Liu, X.~Huang, A.~Mallya, T.~Karras, T.~Aila, J.~Lehtinen, J.~Kautz,
  Few-shot unsupervised image-to-image translation, arXiv preprint
  arXiv:1905.01723.

\bibitem{wang2016relational}
D.~Wang, Y.~Li, Y.~Lin, Y.~Zhuang, Relational knowledge transfer for zero-shot
  learning., in: AAAI, Vol.~2, 2016, p.~7.

\bibitem{socher2013zero}
R.~Socher, M.~Ganjoo, C.~D. Manning, A.~Ng, Zero-shot learning through
  cross-modal transfer, in: Advances in neural information processing systems,
  2013, pp. 935--943.

\bibitem{wang2018zero}
W.~Wang, Y.~Pu, V.~K. Verma, K.~Fan, Y.~Zhang, C.~Chen, P.~Rai, L.~Carin,
  Zero-shot learning via class-conditioned deep generative models, in:
  Thirty-Second AAAI Conference on Artificial Intelligence, 2018.

\bibitem{kingma2013auto}
D.~P. Kingma, M.~Welling, Auto-encoding variational bayes, arXiv preprint
  arXiv:1312.6114.

\bibitem{bucher2017generating}
M.~Bucher, S.~Herbin, F.~Jurie, Generating visual representations for zero-shot
  classification, in: International Conference on Computer Vision (ICCV)
  Workshops: TASK-CV: Transferring and Adapting Source Knowledge in Computer
  Vision, 2017.

\bibitem{xian2018feature}
Y.~Xian, T.~Lorenz, B.~Schiele, Z.~Akata, Feature generating networks for
  zero-shot learning, in: Proceedings of the IEEE conference on computer vision
  and pattern recognition, 2018.

\bibitem{felix2018multi}
R.~Felix, V.~B. Kumar, I.~Reid, G.~Carneiro, Multi-modal cycle-consistent
  generalized zero-shot learning, in: Proceedings of the European Conference on
  Computer Vision (ECCV), 2018, pp. 21--37.

\bibitem{NIPS2013_5027}
R.~Socher, M.~Ganjoo, C.~D. Manning, A.~Ng, Zero-shot learning through
  cross-modal transfer, in: C.~J.~C. Burges, L.~Bottou, M.~Welling,
  Z.~Ghahramani, K.~Q. Weinberger (Eds.), Advances in Neural Information
  Processing Systems 26, Curran Associates, Inc., 2013, pp. 935--943.

\bibitem{chao2016empirical}
W.~Chao, S.~Changpinyo, B.~Gong, F.~Sha, An empirical study and analysis of
  generalized zero-shot learning for object recognition in the wild, in:
  Computer Vision - {ECCV} 2016 - 14th European Conference, Amsterdam, The
  Netherlands, October 11-14, 2016, Proceedings, Part {II}, 2016, pp. 52--68.
\newblock \href {http://dx.doi.org/10.1007/978-3-319-46475-6\_4}
  {\path{doi:10.1007/978-3-319-46475-6\_4}}.

\bibitem{chen2016infogan}
X.~Chen, Y.~Duan, R.~Houthooft, J.~Schulman, I.~Sutskever, P.~Abbeel, Infogan:
  Interpretable representation learning by information maximizing generative
  adversarial nets, in: Advances in neural information processing systems,
  2016, pp. 2172--2180.

\bibitem{he2016deep}
K.~He, X.~Zhang, S.~Ren, J.~Sun, Deep residual learning for image recognition,
  in: Proceedings of the IEEE conference on computer vision and pattern
  recognition, 2016, pp. 770--778.

\bibitem{kingma2014adam}
D.~Kingma, J.~Ba, Adam: A method for stochastic optimization, arXiv preprint
  arXiv:1412.6980.

\bibitem{reed2016learning}
S.~Reed, Z.~Akata, H.~Lee, B.~Schiele, Learning deep representations of
  fine-grained visual descriptions, in: Proceedings of the IEEE Conference on
  Computer Vision and Pattern Recognition, 2016, pp. 49--58.

\bibitem{NIPS2017_7240}
M.~Heusel, H.~Ramsauer, T.~Unterthiner, B.~Nessler, S.~Hochreiter,
  \href{http://papers.nips.cc/paper/7240-gans-trained-by-a-two-time-scale-update-rule-converge-to-a-local-nash-equilibrium.pdf}{Gans
  trained by a two time-scale update rule converge to a local nash
  equilibrium}, in: I.~Guyon, U.~V. Luxburg, S.~Bengio, H.~Wallach, R.~Fergus,
  S.~Vishwanathan, R.~Garnett (Eds.), Advances in Neural Information Processing
  Systems 30, Curran Associates, Inc., 2017, pp. 6626--6637.
\newline\urlprefix\url{http://papers.nips.cc/paper/7240-gans-trained-by-a-two-time-scale-update-rule-converge-to-a-local-nash-equilibrium.pdf}

\bibitem{akata2015evaluation}
Z.~Akata, S.~Reed, D.~Walter, H.~Lee, B.~Schiele, Evaluation of output
  embeddings for fine-grained image classification, in: Proceedings of the IEEE
  Conference on Computer Vision and Pattern Recognition, 2015, pp. 2927--2936.

\bibitem{romera2015embarrassingly}
B.~Romera-Paredes, P.~Torr, An embarrassingly simple approach to zero-shot
  learning, in: International Conference on Machine Learning, 2015, pp.
  2152--2161.

\bibitem{maaten2008visualizing}
L.~v.~d. Maaten, G.~Hinton, Visualizing data using t-sne, Journal of machine
  learning research 9~(Nov) (2008) 2579--2605.

\end{thebibliography}

\end{document}